**Research on Graph-Retrieval Augmented Generation Based on Historical Text Knowledge Graphs**

Yang Fan[1], Zhang Qi[2], Xing Wenqian[1], Liu Chang[1] and Liu Liu[1,*]

[1]College of Information Management, Nanjing Agricultural University, Nanjing 210095, People's Republic of China

[2]School of Economics and Management, Shanxi University, Taiyuan 030006, People's Republic of China

*Correspondence author. E-mail: liuliu@njau.edu.cn

**Abstract**

This article addresses domain knowledge gaps in general large language models for historical text analysis in the context of computational humanities and AIGC technology. We propose the Graph RAG framework, combining chain-of-thought prompting, self-instruction generation, and process supervision to create a "*The First Four Histories*" character relationship dataset with minimal manual annotation. This dataset supports automated historical knowledge extraction, reducing labor costs. In the graph-augmented generation phase, we introduce a collaborative mechanism between knowledge graphs and retrieval-augmented generation, improving the alignment of general models with historical knowledge. Experiments show that the domain-specific model Xunzi-Qwen1.5-14B, with Simplified Chinese input and chain-of-thought prompting, achieves optimal performance in relation extraction (F1 = 0.68). The DeepSeek model integrated with GraphRAG improves F1 by 11% (0.08 → 0.19) on the open-domain C-CLUE relation extraction dataset, surpassing the F1 value of Xunzi-Qwen1.5-14B (0.12), effectively alleviating "hallucinations" phenomenon, and improving interpretability. This framework offers a low-resource solution for classical text knowledge extraction, advancing historical knowledge services and humanities research.

**Keywords:** GraphRAG; Knowledge graph; Knowledge extraction; Process supervision; Large language model

# 1. Introduction

In recent years, the deep integration of Digital Humanities (DH) and Artificial Intelligence Generated Content (AIGC) technologies has revolutionized traditional paradigms in ancient text research. As an interdisciplinary field, DH leverages data mining, corpus construction, and visualization techniques to transcend the subjective limitations of conventional humanities studies, providing objective data support for humanistic research. Within the DH technological framework, Large Language Models (LLMs) like ChatGPT, with their strengths in natural language processing, have opened new avenues for the intelligent processing of ancient texts.

Historical texts, as significant ancient documents, profoundly reflect the social development of the Chinese nation. The core of historical characters knowledge graphs lies in structuring and linking historical figures and their complex relationships, addressing the "information silo" problem caused by fragmented information dispersed across various sources. By employing entity recognition, relationship extraction, and visualization techniques, fragmented data on individuals' biographies, social connections, and historical events can be integrated into a unified network, constructing a multidimensional associative network that reveals latent connections often overlooked in traditional research (Zhang et al. 2024). Simultaneously, LLMs, trained on extensive datasets, can learn from unlabeled data, covering diverse knowledge topics and handling complex tasks (Huang et al. 2025). However, due to the intricate relationships among historical figures, general-purpose LLMs, lacking specialized training, may struggle to accurately identify low-frequency entities and complex semantic associations in ancient texts, leading to historical "hallucinations" and limiting their application in high-precision tasks like historical characters relationship mining (Tonmoy et al. 2024).

As the demand for knowledge extraction from ancient texts deepens, corresponding technologies have evolved through three stages: rule-based matching, machine learning, and deep learning. Early rule-based methods, relying on manual templates, had limited generalization (Zhu et al. 2016; Liang 2021); machine learning introduced feature engineering to enhance adaptability but remained constrained in low-resource scenarios (Tang 2013; Ma and Feng 2024); deep learning methods achieved automation through implicit feature learning (Zhang et al. 2022; Liu et al. 2023), The new paradigm driven by LLMs further breaks through local optimization limitations, demonstrating cross-corpus generalization potential (Liu et al. 2025). However, current research faces two main challenges: the absence of a cross-historical characters relationship corpus, hindering multi-source knowledge integration, and the adaptability of large models in low-resource ancient text tasks, with insufficient collaborative validation with downstream applications.

To address these challenges, Retrieval-Augmented Generation (RAG) technology offers a promising solution. RAG enhances the domain capabilities of general-purpose LLMs cost-effectively by integrating external knowledge retrieval, suppressing "hallucinations" and improving response credibility (Lewis et al. 2020). Its core comprises two key components: retrieval and generation. Retrieval focuses on semantic alignment and data optimization, exemplified by datasets like HotpotQA (Yang et al. 2018), frameworks such as PROMPTAGATOR for few-shot query generation (Dai et al. 2022), and graph data retrieval optimization based on Steiner trees (He et al. 2024). Generation integrates retrieved information to enhance output accuracy, with typical studies including the "adaptive filtering-elimination" paradigm (Ma et al. 2022) and the "retrieval-ranking" joint framework (Zhang et al. 2022). However, traditional RAG, relying on unstructured text retrieval, may fail to capture deep semantic associations and complex knowledge inference paths (Gao et al. 2024). In response, GraphRAG utilizes structured knowledge graphs to manifest explicit associations, employing methods like the Neo4j graph query framework (Cheng et al. 2025; Peng 2025) for precise knowledge localization and vector space mapping to enhance edge knowledge recall, as seen in cross-document semantic indexing (Wang et al., 2023) and subgraph contrastive learning (Kang et al. 2023). Studies indicate that GraphRAG, through multi-hop reasoning and explicit edge relationships, significantly enhances the interpretability of complex issues, though its application in ancient text research remains limited.

Building upon this research context, this study investigates graph retrieval-augmented generation based on historical characters knowledge graphs. The research comprises two main parts: first, knowledge graph construction, where we develop a characters relationship ontology model and corpus covering the "*The First Four Histories*" Through techniques such as chain-of-thought prompting, self-instruction, and process supervision, we achieve high-precision automated knowledge extraction from historical texts with minimal manual intervention, providing a reusable technical pathway for constructing ancient text knowledge graphs. Second, graph retrieval-augmented generation, where we deeply integrate historical characters knowledge graphs with RAG technology, exploring the synergistic mechanism of graph-based and semantic retrieval in the historical GraphRAG framework. This approach aims to provide high-quality background information for complex issues and evaluates the enhancement of domain capabilities in general-purpose LLMs within the historical GraphRAG framework. This study aims to offer an interpretable and cost-effective technical paradigm for the digital preservation of ancient texts and humanistic research, promoting the digital transformation of historical knowledge reconstruction.

The paper is structured as follows: Section 2 introduces the overall design framework of the study; Section 3 presents a multi-dimensional evaluation of the efficacy of large models in ancient text knowledge extraction and details the automated extraction process; Section 4, based on the constructed historical characters knowledge graph, develops the historical GraphRAG system and tests its performance; Section 5 discusses and concludes the research.

## 2. Research Design

*2.1 Research Plan*

This study is divided into two main components: the construction of the knowledge graph and GraphRAG. The knowledge graph construction spans the ontology and data layers, as well as the instance layer, focusing on defining historical figures' ontologies, conducting performance evaluation experiments, and generating instantiated knowledge networks. The GraphRAG module, built upon the structured knowledge from the instance layer, implements semantic retrieval and generative knowledge services at the application layer. The research framework is detailed in Figure 1:

①Ontology and Data Layer: By reusing historical ontology models to construct the historical figures' relationship ontology, and combining minimal manual annotations with large language model self-instruction techniques, we developed datasets for chain-of-thought reasoning and few-shot prompting strategies. ②Instance Layer: Through comparative validation of the effectiveness of chain-of-thought prompting strategies, simplified chinese input and domain-specific fine-tuning of the model, we constructed an "automated extraction-scoring" dual-model system based on the Xunzi-Qwen1.5-14B model, achieving a full-process intelligent construction of the "*The First Four Histories*" character relationship corpus. ③Application Layer: Utilizing DeepSeek as both the retriever and generator, we designed the GraphRAG architecture, integrating Neo4j graph storage, DeepSeek's semantic parsing, and generative enhancement technologies. This integration employs a graph-structure and semantic collaborative retrieval method and knowledge-enhanced generation to enhance the domain-specific knowledge reasoning capabilities of general-purpose LLMs.

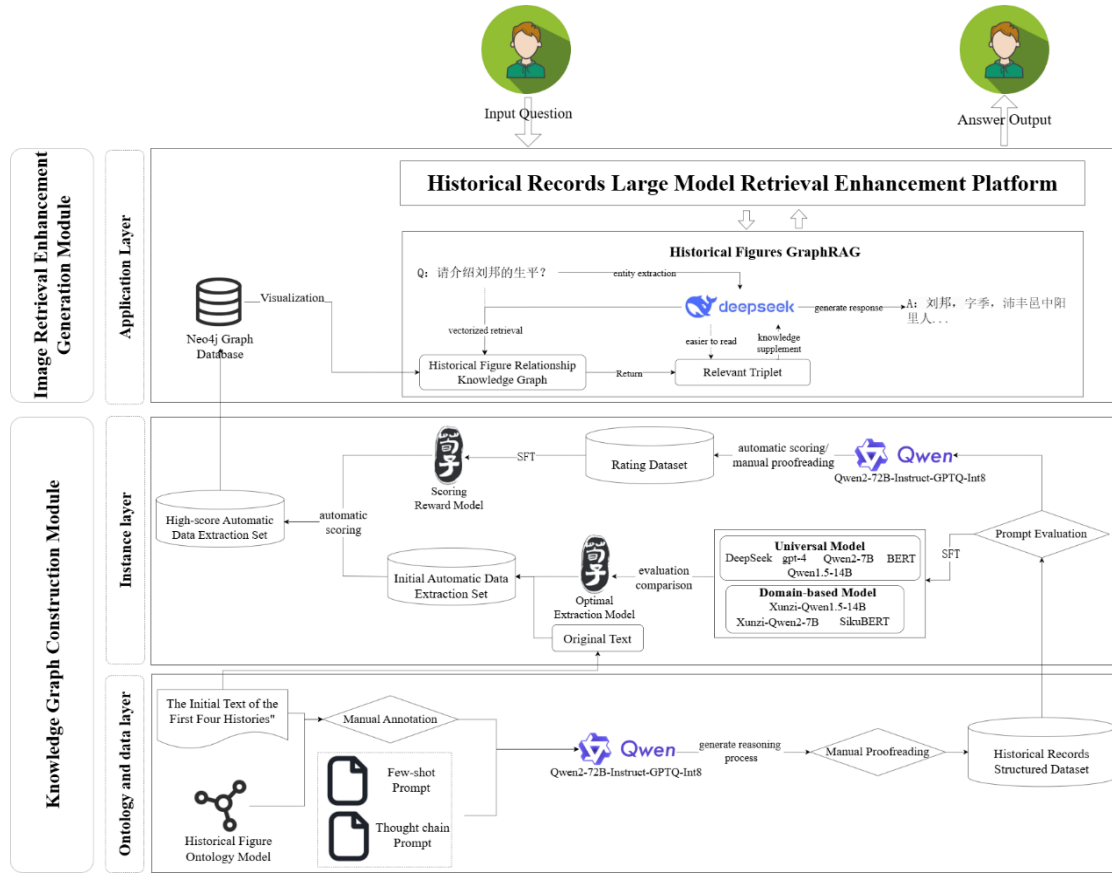

Figure 1. Research Framework

## 2.2 Domain Oriented Ontology And Dataset Construction

### 2.2.1 Domain Oriented Ontology Construction

This study is based on the multi-dimensional knowledge description ontology model of historical texts (Zhang et al. 2022). By surveying online historical ontology model databases, we extracted conceptual terms related to personal relationships and reconstructed the classification system to develop an ontology model for historical figures in classical texts. The model encompasses six entity types: Person, Location, Official Position, Noble Title, And Social Organization. Detailed conceptual relationships and supplementary explanations are provided in Figure 2.

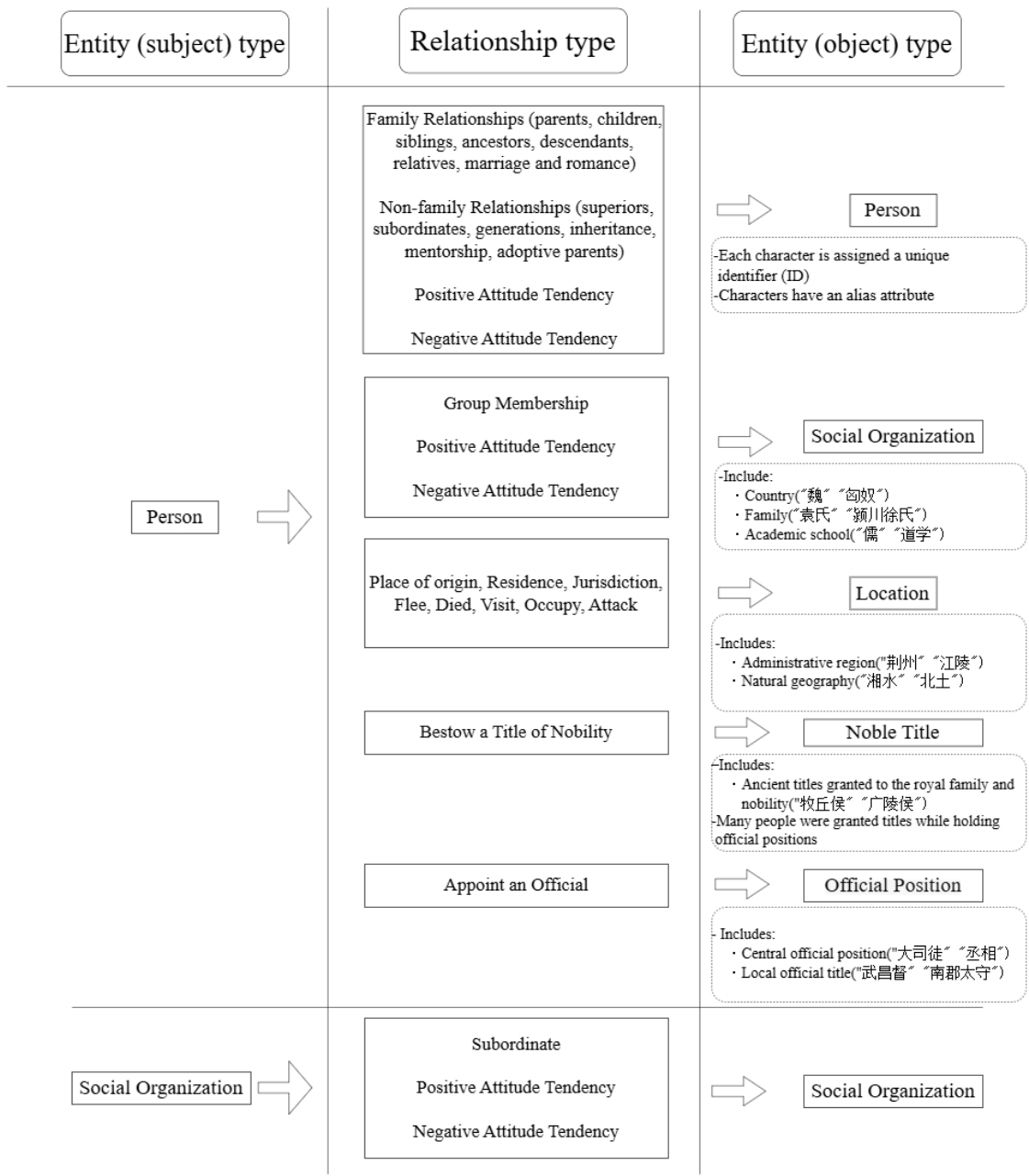

Figure 2. Modeling of Historical Character Relationships Knowledge

*2.2.2 Dataset Source*

This study utilizes the original texts of the "*The First Four Histories*"—Records of the Grand Historian, Book of Han, Book of the Later Han, and Records of the Three Kingdoms—from the Chinese Text Project (https://ctext.org/zh) as the primary corpus. A total of 5,000 passages were randomly sampled in proportion to the length of each text. Two annotators independently labeled these passages to construct a corpus of character relationship data.

To evaluate model performance variations, a high-quality subset of the Records of the Grand Historian was designated as a control group. Comparative experiments were conducted to assess the impact of different prompting strategies on processing efficacy across historical periods. Detailed information on the historical texts used and sample passages are presented in Table 1 and Figure 3, respectively.

Table 1 Sources of Historical Corpus

| Historical Texts | Number of Sentences | Character Count | Proportion |
|---|---|---|---|
| *Records of the Grand Historian* | 30501 | 607999 | 26% |
| *Book of Han* | 39776 | 881415 | 34% |
| *Book of the Later Han* | 26887 | 607500 | 23% |
| *Records of the Three Kingdoms* | 18756 | 471324 | 16% |

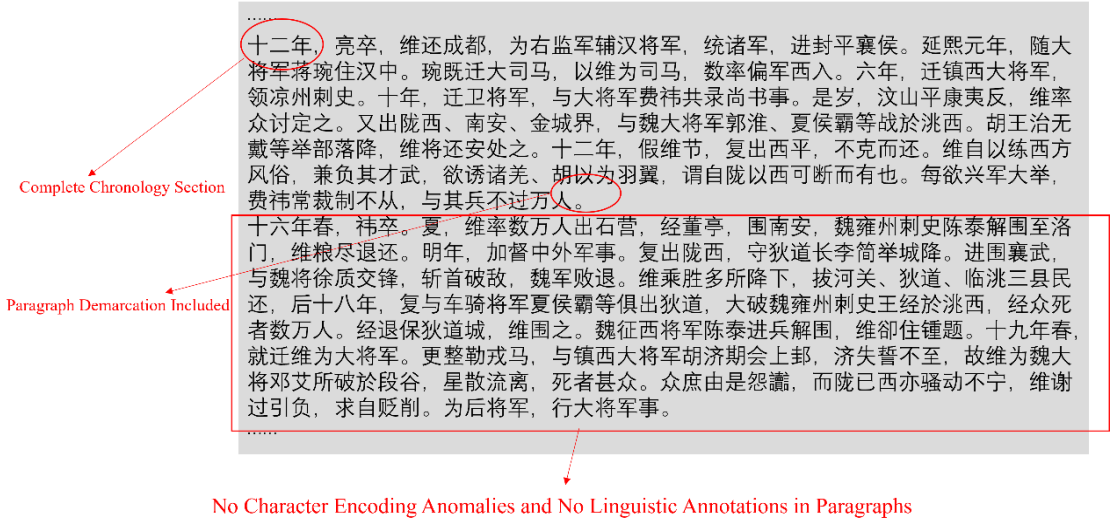

Figure 3. Example of Historical Corpus

### 2.3 Prompt Strategy Design

#### 2.3.1 Design Prompt

Due to the significant resource investment required for fine-tuning large language models, and given that the quality of prompt information directly affects the model's performance on specific tasks, prompt engineering based on natural language instructions has gradually become the preferred strategy for addressing downstream tasks. Large language model prompt engineering involves elements such as role setting, task description, context embedding, example guidance, and output control. This study designs 5-shot and 0-COT prompts based on these elements.

Few-shot prompt engineering is a technique that guides the model to generate task-specific outputs by providing contextual examples to the large language model (Bouniot et al. 2022). Unlike zero-shot learning, few-shot learning establishes task patterns through examples, reducing the model's reliance on large-scale labeled data, while improving output quality for complex tasks. This study constructs few-shot prompts using examples that cover all entities, as shown in Figure 4. Additionally, due to significant differences in the morphological and syntactical rules between Classical and Modern Chinese, extracting relationship triplets from ancient texts is more challenging than from contemporary texts. To enhance comprehension of ancient texts, annotators often need to perform semantic transformation and structural reorganization. This cognitive process, when mapped to the large language model's processing mechanism, can improve performance on complex tasks (Wei et al. 2023). Based on this, the relationship extraction process from historical texts is abstracted into five steps: Translation into Modern Language, Named Entity Recognition, Semantic Relation Analysis, Construction of The Relationship Framework, and Extraction of Entity-Relation Triplets. Chain-of-thought prompts are designed to simulate the cognitive intermediary process of relationship extraction, as detailed in Figure 5.

Figure 4. Design of 5-Shot Prompt

Figure 5. Design of COT Prompt

### 2.3.2 Construct A Chain-Of-Thought Dataset

This study employs Self-Instruct technology to construct a knowledge extraction reasoning dataset (Wang et al. 2023) . By deploying the Qwen2-72B-Instruct model on annotated data from the "*The First Four Histories*" and Records of the Grand Historian, we generated 5,000 structured reasoning chains through chain-of-thought reasoning processes. These datasets were divided into training and testing sets in a 9:1 ratio, providing standardized data support for low-intervention automated knowledge extraction.

Detailed methodologies are described in Figure 6.

```
{
    "instruction": "你是一个古文专家，根据提供的古文抽取实体关系scheme三元组，常见史籍人物关系实体有：[人物,社会集团,地点,时间,职官,爵位]。现在需要你从原文中一步步推理得到该正确关系抽取结果，推理步骤包括：1.翻译为现代文、2.命名实体识别、3.实体间语义关联分析、4.构建关系模式框架、5.实体关系三元组抽取。\n 抽取原则为：1.若无对应的关系scheme组合，则不抽取任何内容，尽可能少而精确得抽取scheme组合；2.抽取时一定要符合限定的scheme框架。\n 抽取格式为：(实体实例1:实体类型,关系类型,实体实例2:实体类型)。\n 输入文本为：",
    "input": "愍王元年,秦使张仪与诸侯执政会于啮桑。",
    "output": "1.翻译为现代文：愍王元年，秦国派遣张仪与各国执政在啮桑会面。\n 2.命名实体识别：张仪（人物）、秦（社会集团）、啮桑（地点）\n 3.实体间语义关联分析：张仪是秦国派遣的使者，代表秦国与各国执政会面，因此张仪属于秦国。张仪与各国执政会面的地点是啮桑。\n 4.构建关系模式框架：(人物,所属团体,社会集团)，(人物,访,地点)\n 5.实体关系三元组抽取：(张仪:人物,所属团体,秦:社会集团)，(张仪:人物,访,啮桑:地点)\n"
}
```

Figure 6. Structure Illustration of the Instruction Dataset

*2.4 Base Model Selection*

*2.4.1 General-Purpose Language Models*

For general-purpose models, we assess performance using API calls to billion-parameter models, specifically GPT-4 and DeepSeek. DeepSeek, noted for its high inference efficiency and cost-effectiveness, serves as a core component of GraphRAG. We conduct horizontal comparisons with the Qwen series models (7B and 14B parameters) to evaluate the impact of parameter scale and architectural optimizations on tasks involving ancient texts. Additionally, we introduce BERT-based models as traditional deep learning baselines to comprehensively assess the applicability boundaries of generative and discriminative models in relationship extraction tasks (Devlin et al. 2019).

*2.4.2 Domain-Specific Language Models for Ancient Texts*

For domain-specific models, this study integrates the "Xunzi" series models developed by Nanjing Agricultural University (Wang et al. 2024), with parameter scales of 7B and 14B. We compare the effects of different parameter sizes and training paradigms on relationship extraction performance. Furthermore, we incorporate SikuBERT, a BERT-based model (WANG et al. 2022), as a baseline for domain-specific pre-trained models, establishing a horizontal performance comparison between generative and pre-trained models. Detailed functionalities of each model are presented in Table 2.

Table 2. Model Capabilities

| Model Name | Model Type | Prompt Strategy and Input Glyph Evaluation | Model Performance Evaluation | Automated Knowledge Extraction & Scoring Reward Model | GraphRAG Retriever & Generator |
|---|---|---|---|---|---|
| GPT-4 | General Language Models | | ✓ | | |
| DeepSeek | | | ✓ | | ✓ |
| Qwen2-7B | | ✓ | ✓ | | |
| Qwen1.5-14B | | | ✓ | | |
| BERT | | | ✓ | | |
| Xunzi-Qwen2-7B | Classical Text-Specific | ✓ | ✓ | | |
| Xunzi-Qwen1.5-14B | | | ✓ | ✓ | |

| | SikuBERT | Language Models | | ✓ | | |

### 2.5 Evaluation Metric Selection

#### 2.5.1 F1

F1 is the harmonic mean of precision and recall, commonly used to evaluate the performance of classification tasks (Paroubek et al. 2007). In relation extraction tasks, F1 provides a comprehensive measure of the model's accuracy and completeness in identifying entity relationships. The specific formula for calculating F1 is as follows:

$$\text{Precision} = \frac{\text{TP}}{\text{TP} + \text{FP}} \tag{1}$$

$$\text{Recall} = \frac{\text{TP}}{\text{TP} + \text{FN}} \tag{2}$$

$$\text{F1-Score} = \frac{2 \times \text{precision} \times \text{recall}}{\text{precision} + \text{recall}} \tag{3}$$

#### 2.5.2 ROUGE

ROUGE Metrics inclue: ROUGE-1, ROUGE-2, and ROUGE-L are selected to compute the text overlap rate to assess the similarity between generated and reference texts (Lin 2004). In relation extraction tasks, ROUGE metrics are often used to measure the lexical matching between the model-generated relation triples and the reference. The specific formulas for calculating ROUGE are as follows:

$$ROUGE - 1 = \frac{Number\ of\ overlapping\ unigrams}{Total\ number\ of\ unigrams\ in\ reference} \tag{4}$$

$$ROUGE - 2 = \frac{Number\ of\ overlapping\ unigrams}{Total\ number\ of\ unigrams\ in\ reference} \tag{5}$$

$$ROUGE - L = \frac{LCS(Generated, Reference)}{Length\ of\ Reference} \tag{6}$$

### 2.6 Experimental Parameter Configuration

This study utilizes the PyTorch-2.4.0 deep learning framework and the transformers-4.45.0 library for model deployment, with a dual A800 GPU cluster and CUDA 12.2 acceleration for computational efficiency. To accommodate long-text semantic input, an input length of 4096 characters is set. Additionally, a learning rate warm-up strategy is employed to enhance model convergence speed. Key hyperparameter configurations are detailed in Table 3.

Table 3. Key Hyperparameters for Model Fine-tuning

| Parameter Name | Parameter Description | Value |
|---|---|---|
| epoch | Total number of training iterations | 3.0 |
| per_device_train_batch_size | Training batch size per GPU device | 4.0 |
| Max_length | Maximum token length for input sequences | 4096 |
| learning_rate | Optimization step size for gradient descent | 5e-5 |
| Temperature | Coefficient controlling output randomness | 0 |

## 3. Historical Figure Knowledge Graph Construction

*3.1 Multidimensional Evaluation of Large Language Models for Knowledge Extraction from Classical Texts*

This study constructs a multidimensional evaluation framework, systematically quantifying the performance differences of prompt strategies (few-shot/chaining), Glyph Input (Simplified and Traditional Chinese Characters) and model types (general-purpose vs. domain-specific) by controlling these three variables. It defines the capability boundaries of general-purpose and domain-specific models in historical document tasks and integrates the mean values of four metrics to generate a comprehensive SCORE, enabling cross-dimensional analysis of model performance.

*3.1.1 Prompt Strategy Evaluation*

To investigate the impact of prompt type differences on the extraction performance of large language models, this study fine-tuned and evaluated models using 4000 samples of Simplified Chinese glyphs from the "*The First Four Histories*" and the research group's *Records of the Grand Historian* corpus, including few-shot and chain-of-thought instruction data. The evaluation results are presented in Table 4. Under the chain-of-thought prompt scenario, all models showed a significant improvement in F1 scores compared to few-shot prompts, especially in the "*The First Four Histories*" resource, where the F1 (0.60) and SCORE (0.75) of Xunzi reached the highest values. Therefore, subsequent experiments adopted the chain-of-thought strategy. Furthermore, models performed better with the "*The First Four Histories*" resource compared to the Records of the Grand Historian, indicating that the self-constructed "*The First Four Histories*" corpus has higher quality and is suitable for subsequent automated knowledge extraction tasks.

Table 4. Performance Evaluation of Prompt-Engineered

| Model Name | Category Description | F1 | ROUGE-1 | ROUGE-2 | ROUGE-L | SCORE |
|---|---|---|---|---|---|---|
| Xunzi-Qwen2-7B | "*The First Four Histories*"+Strategy of 0-COT Prompt | **0.60** | 0.85 | 0.72 | 0.82 | **0.75** |
| Qwen2-7B | | 0.48 | 0.85 | 0.72 | 0.82 | 0.72 |
| Xunzi-Qwen2-7B | "*The First Four Histories*"+Strategy of 5-Shot Prompt | 0.49 | 0.81 | 0.68 | 0.77 | 0.69 |
| Qwen2-7B | | 0.47 | 0.80 | 0.67 | 0.77 | 0.68 |
| Xunzi-Qwen2-7B | *Records of the Grand Historian*+Strategy of 0-COT Prompt | 0.55 | 0.85 | 0.73 | 0.83 | 0.74 |
| Qwen2-7B | | 0.49 | 0.85 | 0.73 | 0.83 | 0.72 |
| Xunzi-Qwen2-7B | *Records of the Grand Historian*+ Strategy of 5-Shot Prompt | 0.53 | 0.84 | 0.73 | 0.82 | 0.73 |
| Qwen2-7B | | 0.48 | 0.84 | 0.73 | 0.82 | 0.72 |

*3.1.2 Input Glyph Evaluation*

Simplified Chinese characters were introduced in Mainland China during the 1950s as part of a character simplification system, aiming to improve writing efficiency through stroke reduction, structural simplification (e.g., "语" derived from "語"), and homophone merging. The core goal was to lower literacy barriers and enhance practicality. In contrast, Traditional Chinese characters preserve the original structure. They are more complex in form and carry richer historical and cultural information, with stronger logical consistency in character formation. These two forms of writing differ significantly in terms of structural complexity, usage regions, and cultural attributes, representing functional and symbolic variants within the Chinese character system.

To investigate the impact of glyph input on the performance of large language models, this study extracted a 3000-sample chain-of-thought dataset from the "*Records of the Grand Historian*" corpus. The models Xunzi-Qwen2-7B and Qwen2-7B were fine-tuned with Traditional and Simplified Chinese glyph inputs, respectively. The experimental results, as shown in Table 5, reveal that the Xunzi-Qwen2-7B model with Simplified Chinese input outperforms others. Specifically, Xunzi-Qwen2-7B shows a 15% increase in F1 score with Simplified Chinese input, while Qwen2-7B's F1 score improves by 25%. The performance gap between the two models narrows under Simplified Chinese input (with only a 1% F1 difference), and some metrics even show parity between the models.

In contrast, Xunzi-Qwen2-7B performs better with Traditional Chinese input, likely benefiting from continued pretraining in the ancient text domain. Overall, the results suggest that models with Simplified Chinese input perform better than those with Traditional Chinese input. This indicates that model performance is sensitive to the input glyphs and relates to the adaptability of the training corpus. Potential causes include the higher coverage of Simplified Chinese characters in the training corpus and semantic unit segmentation biases caused by differences in Traditional Chinese encoding patterns. Therefore, to automate the extraction of historical knowledge from original texts, Simplified Chinese glyphs are employed in historical document experiments.

Table 5. Performance Evaluation of Input Models for Traditional and Simplified Chinese Characters

| Model Name | Input Glyph | F1 | ROUGE-1 | ROUGE-2 | ROUGE-L | SCORE |
|---|---|---|---|---|---|---|
| Xunzi-Qwen2-7B | Simplified Chinese Characters | **0.55** | 0.93 | 0.87 | 0.90 | **0.81** |
| Qwen2-7B | | 0.54 | 0.93 | 0.87 | 0.90 | 0.80 |
| Xunzi-Qwen2-7B | Traditional Chinese Characters | 0.40 | 0.90 | 0.82 | 0.88 | 0.75 |
| Qwen2-7B | | 0.29 | 0.88 | 0.77 | 0.85 | 0.70 |

*3.1.3 Performance Evaluation of Extraction Models*

Given the complexity of domain-specific relation extraction tasks, this study enhances model stability and accuracy by performing instruction fine-tuning (SFT) on open-source models. Additionally, 5-shot prompt examples are incorporated to improve extraction precision. Both chain-of-thought prompting and Simplified Chinese input, previously shown to yield superior performance, are employed across all models; detailed evaluation results are presented in Table 6.

Horizontally, the Xunzi-Qwen1.5-14B model exhibits significant advantages across all metrics. Its F1 score reaches 0.68, markedly surpassing other models. High scores in ROUGE-1 (0.97), ROUGE-2 (0.94), and ROUGE-L (0.96) indicate strong alignment with reference triples in terms of vocabulary, syntax, and semantics. In contrast, general models like GPT-4o and DeepSeek achieve F1 scores of 0.28 and 0.23, respectively, highlighting limited adaptability to classical texts without domain-specific training.

Vertically, significant performance disparities are observed within models across the same metric. Xunzi series models generally outperform general models in F1 scores. Notably, smaller pre-trained models rival certain large language models (LLMs), demonstrating value in low-resource scenarios. Xunzi-Qwen1.5-14B leads with an F1 score of 0.68, nearly 4% higher than Xunzi-Qwen2-7B's 0.64, underscoring the importance of parameter scale. In ROUGE metrics, Xunzi series models consistently outperform their base models, especially ROUGE-1, reflecting the benefits of domain adaptation.

Specifically, Xunzi-Qwen1.5-14B excels in domain adaptability and parameter scale, achieving superior understanding of classical Chinese semantics through pre-training on classical corpora and instruction fine-tuning for relation extraction tasks.

In summary, domain-adapted open-source models offer significant advantages in classical Chinese semantic understanding. While general-purpose open-source models provide strong generalization and extensive ecosystem support, they lack adaptability to classical Chinese and may produce lower-quality outputs. Proprietary models excel in inference speed and deployment convenience but often lack optimization for classical texts. Therefore, for automated knowledge extraction from historical texts, Xunzi-Qwen1.5-14B is recommended as the optimal model.

Table 6. Comparative Performance of General-Purpose and Domain-Specialized Models on Constrained Relation Extraction from Classical Texts

| Model Name | Processing Method | F1 | ROUGE-1 | ROUGE-2 | ROUGE-L | SCORE |
| --- | --- | --- | --- | --- | --- | --- |
| GPT-4o | 5-Shot | 0.28 | 0.74 | 0.54 | 0.71 | 0.57 |
| DeepSeek | 5-Shot | 0.23 | 0.75 | 0.60 | 0.71 | 0.57 |
| Qwen1.5-7B | SFT | 0.51 | 0.93 | 0.87 | 0.92 | 0.81 |
| Xunzi-Qwen2-7B | SFT | 0.55 | 0.95 | 0.92 | 0.94 | 0.84 |
| Qwen2-7B | SFT | 0.47 | 0.94 | 0.90 | 0.93 | 0.81 |
| Xunzi-Qwen1.5-14B | SFT | **0.68** | 0.97 | 0.94 | 0.96 | **0.89** |
| Qwen1.5-14B | SFT | 0.48 | 0.94 | 0.90 | 0.94 | 0.82 |
| sikubert | | 0.47 | | | | 0.47 |
| bert | | 0.36 | | | | 0.36 |

*3.2 Formulation of Chain-of-Thought Scoring Reward Rules*

Process supervision is a method of supervision for large model training and inference, focusing on monitoring and rewarding each step of the model's reasoning process rather than only evaluating the final result. By guiding the model to follow the correct logical reasoning path, it aims to enhance the accuracy and interpretability of the responses (Luo et al. 2024). In relation extraction tasks, the scoring supervision rule emphasizes monitoring the reasoning process, evaluating and scoring intermediate results to ensure that each reasoning step aligns with logic and factual accuracy. The specific scoring process includes five key steps: Translation into Modern Language, Named Entity Recognition, Semantic Relation Analysis, Construction of The Relationship Framework, and Extraction of Entity-Relation Triplets (Liu et al. 2025). Scoring is done using a penalty-based system, with each step assigned a maximum score of 2 points, for a total score of 10 points. Detailed scoring criteria are provided in Table 7.

Table 7. Supervised Scoring Rules for Constrained-Domain Relation Extraction Across Reasoning Stages

| Scoring Hierarchy | Deduction Basis | Deduction Points |
| --- | --- | --- |
| Translate into Modern Language | If the translation contains errors | 1 |
| | If the translation is completely unrelated to the original text | 2 |
| Named Entity Recognition | If the identified entity does not appear in the original text | 1 |
| | If the entity information is incomplete | 1 |

| Semantic Relationship Analysis | If the analysis results contain content inconsistent with the original meaning | 1 |
|---|---|---|
| | If errors occur when determining facts | 2 |
| Construction of The Relationship Framework | If there are duplicated relationship schemas (schema) | 1 |
| | If the schema contains meaningless relationship predicates | 1 |
| | If a non-specified relationship schema is used | 1 |
| Extraction of Entity-Relation Triplets | If the extracted triples are incomplete | 1 |
| | If incorrect extraction occurs | 1 |
| | If the extracted triple elements do not appear in the original text | 1 |
| | If the triples are not extracted according to the specified relationship schema | 1 |

This study utilizes a 500-sample evaluation dataset and deploys the Qwen2-72B-Instruct-GPTQ-Int8 model via the vllm tool to establish a five-stage scoring framework. High-quality annotated data are generated through manual validation (examples shown in Table 8). To balance computational efficiency and domain adaptability, the Xunzi-Qwen1.5-14B model, fine-tuned for ancient texts, is employed as the scoring reward model, acting as an automated data filter for triplet extraction.

Table 8. Exemplar Stepwise Scoring Process

| Scoring Hierarchy | Relation Extraction Process | Rationale for Scoring | Score |
|---|---|---|---|
| Translate into Modern Language | 1.翻译为现代文：武王娶了魏国的女子作为王后，但没有生下孩子。 | The translation is accurate and error-free, maintaining complete consistency with the original text. | 2 points |
| Named Entity Recognition | 2.命名实体识别：武王（人物），魏（社会集团） | Entity recognition is executed with precision. The entity types for "武王" and "魏" are correctly categorized in accordance with the source text. | 2 points |
| Semantic Relationship Analysis | 3.实体间语义关联分析：武王对魏国有积极的态度倾向。 | While the statement "武王" married "魏女" as his queen" could be interpreted as implying a positive attitude toward "魏", the original text does not explicitly express this attitudinal tendency. Consequently, a 1-point deduction is applied. | 1 points |
| Construct Relation | 4.构建关系模式框架： | Although a relational schema | 1 points |
| Schema Framework | (人物,积极态度倾向,社会集团) | framework is generated, the inclusion of "积极态度倾向" lacks direct textual evidence in the source material. A 1-point deduction is therefore warranted. | |
| Extraction of Entity-Relation Triplets | 5.实体关系三元组抽取：(武王:人物,积极态度倾向,魏:社会集团) | The extracted triplets demonstrate inconsistencies with the original contextual relationships, particularly regarding the absence of explicit "积极态度倾向" expressions. This discrepancy results in an additional 1-point deduction. | 1 points |
| Total Score | | | 7 points |

*3.3 Automated Relation Extraction Implementation*

This study constructs an automated pipeline based on the optimal extraction model and the scoring reward model. The Xunzi-Qwen1.5-14B model generates an initial dataset of character relation triplets through step-by-step reasoning, followed by dynamic scoring and filtering using the reward model.

The results show that 69.79% of the data received scores above 6 (Figure 7), with scores between 8 and 9 reflecting some triplet omissions but still meeting overall accuracy standards. High-quality data with scores ranging from 8 to 10 were selected as the core corpus, validating the effectiveness of the fully automated knowledge extraction process and the reliability of the data filtering mechanism.

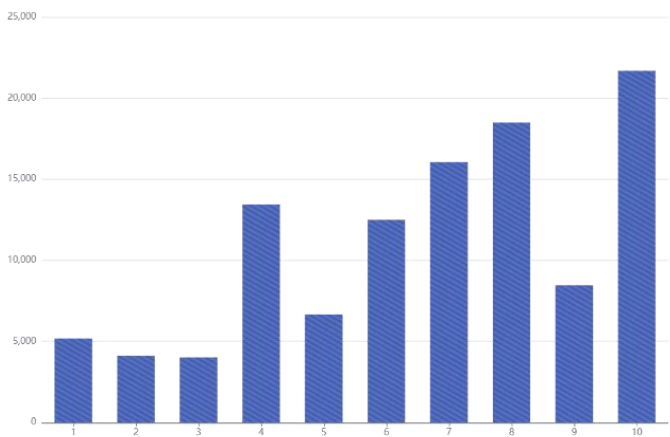

Figure 7. Score Distribution Histogram for Knowledge Extraction Results

Subsequently, this study constructs a historical figure knowledge graph (Neo4j) based on relation triplets and implements multi-dimensional entity association visualization through Cypher queries (e.g., the social network of "Confucius" shown in Figure 8). A structured alias dictionary for figures is also created (e.g., primary name: alias1, alias2…), and combined with data from Han Studies Source Series to form a composite index. This is integrated into the GraphRAG framework to enhance entity linking. Alias expansion effectively improves entity retrieval coverage, mitigating the issue of entity "misses".

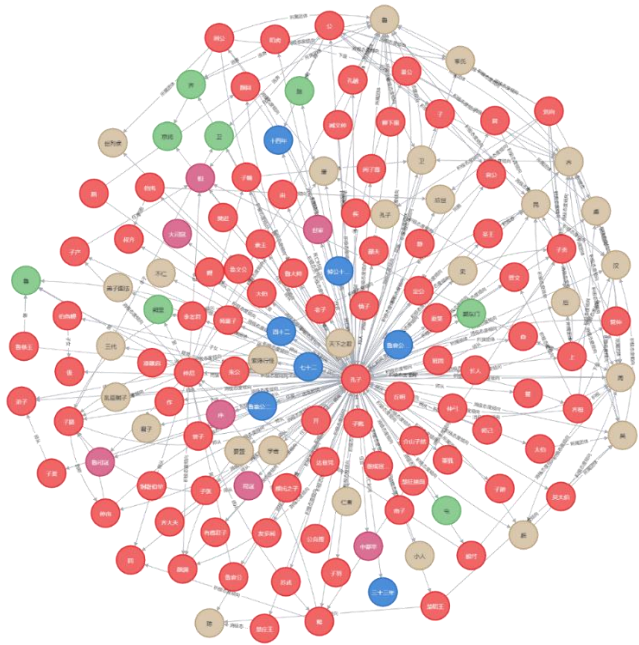

Figure 8. Exemplary Visualization of Historical Figure Knowledge Graph

## 4. Application of Graph-based Retrieval-Augmented Generation for Historical Text

*4.1 Graph-based Retrieval-Augmented Generation for Large Language Models*

This research is based on the previously constructed historical figure knowledge graph and systematically builds a GraphRAG aimed at enhancing the historical knowledge capabilities of general-purpose large models. Specifically, DeepSeek is chosen as the core large model for the GraphRAG system. With its scale of hundreds of billions of parameters, DeepSeek demonstrates significant advantages in inference speed for real-time interaction scenarios. Compared to other general models, its API interface is more stable and user-friendly, balancing cost and performance (DeepSeek-A et al. 2024). The retrieval-augmented generation (RAG) module of this research is developed using the LangChain framework, enabling efficient integration between large language models and external data sources while supporting complex task chains. By utilizing LangChain's GraphCypherQAChain, the user's natural language queries are converted into Cypher query statements, which are executed in the Neo4j graph database to retrieve relevant contextual information. Specifically, the historical figure relationship GraphRAG system is composed of three components: Knowledge Storage, Retriever, and Generator. These components work in tandem to ensure the system efficiently extracts, stores, retrieves, and generates accurate and coherent answers from historical texts, as illustrated in Figure 9. The following outlines the detailed process.

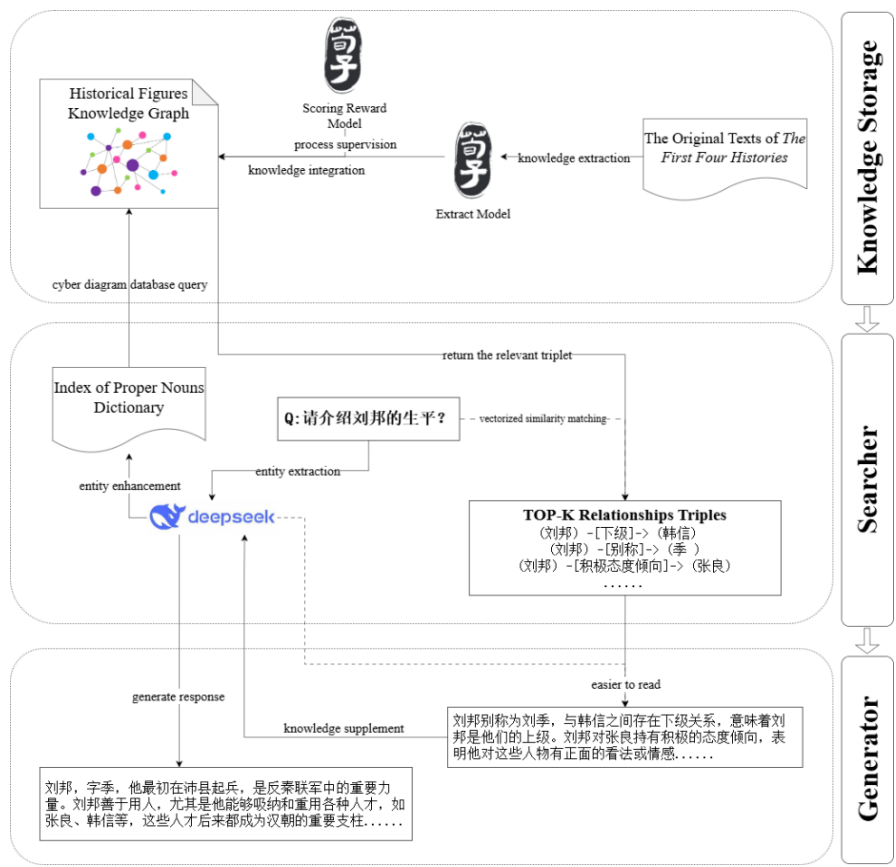

Figure 9. System Workflow for Graph-based Retrieval-Augmented Generation with Historical Figure Context

**(1) Knowledge Storage:** Responsible for storing knowledge-integrated historical figure relationship data in the graph database to support subsequent retrieval and generation tasks. Specifically, the system utilizes the extraction model to automatically extract relation triplets from the original texts of the "*The First Four Histories*" and filters high-quality triplet data through the scoring reward model (see Section 3.3). The data are then imported into the Neo4j graph database to generate a visualized knowledge graph. Leveraging the storage structure of the graph database, the system efficiently manages and queries complex entity relationship networks.

**(2) Retriever:** Extracts relevant entities from user queries and retrieves associated triplets from the knowledge graph, selecting results closest in semantic meaning to the user's question. The workflow is as follows: Upon receiving a query, the system employs the DeepSeek model for semantic analysis to extract key entities. To prevent entity omissions, the extracted entities are cross-referenced with the structured alias dictionary, supplementing with primary names or aliases to broaden the retrieval scope. For example, for the query " What are the aliases of Cao Cao?", the system extracts "Cao Cao" as an entity, which is then expanded to include aliases such as "Wei Taizu", "Mengde", "Aman" and "Cao Mengde" Subsequently, the system uses GraphCypherQAChain to convert these entities into Cypher query statements, retrieves relevant triplets from the graph database, and processes them with the paraphrase-multilingual-MiniLM-L12-v2 model for vectorization. Through vector similarity calculations, the top-K most relevant triplets are selected.

**(3) Generator:** Transforms the retrieved triplets into coherent, readable textual responses. The process includes: First, converting structured triplets into natural language text. For instance, a triplet

(Cao Cao:person, children, Cao Pi:person) is transformed into "Cao Cao's child is Cao Pi." enhancing readability and naturalness. Next, the system incorporates the retrieved triplets as prior knowledge into DeepSeek's prompt, formatted as: "You are an expert in classical Chinese literature. Referencing the following information: {readable relation triplets}, based on your knowledge, answer the question: {user query}." If the knowledge in the graph database is insufficient to generate a response, the system replies: "Unable to answer to prevent generating incorrect information." This method allows the generator to produce more accurate and relevant answers based on the retrieved background knowledge, effectively mitigating the "hallucination" problem in large language models.

*4.2 System Performance Evaluation*

This study conducts comparative experiments to validate the optimization effectiveness of GraphRAG in historical document question-answering systems. Comparisons based on the Xunzi-Qwen series models reveal that Xunzi-Qwen1.5-*14B* achieves optimal performance on the *"The First Four Histories"* dataset (F1=0.65, SCORE=0.87), highlighting the synergistic effect of parameter scale and domain-specific fine-tuning. The general-purpose model DeepSeek, after enhancement with GraphRAG, shows a significant performance improvement (F1 increased from 0.21 to 0.34), confirming the efficacy of knowledge alignment and hallucination suppression. In cross-domain testing, DeepSeek+RAG surpasses the native Xunzi-Qwen1.5-14B (F1=0.15) on the C-CLUE dataset (F1=0.19), revealing the technical potential of GraphRAG in extending domain knowledge for general-purpose LLMs. Detailed results are presented in Table 9.

Table 9. Performance Evaluation of Historical Figure Relationship GraphRAG System

|  | Dataset Type | F1 | ROUGE-1 | ROUGE-2 | ROUGE-L | SCORE |
| --- | --- | --- | --- | --- | --- | --- |
| Xunzi-Qwen2-7B | The manually annotated dataset of "*The First Four Histories*" | 0.54 | 0.95 | 0.91 | 0.94 | 0.83 |
| Xunzi-Qwen1.5-14B | | **0.65** | 0.96 | 0.93 | 0.95 | **0.87** |
| DeepSeek | | 0.21 | 0.87 | 0.76 | 0.82 | 0.66 |
| DeepSeek+RAG | | **0.34** | 0.89 | 0.79 | 0.86 | **0.72** |
| Xunzi-Qwen2-7B | C-CLUE | 0.14 | 0.50 | 0.36 | 0.49 | 0.38 |
| Xunzi-Qwen1.5-14B | | **0.12** | 0.53 | 0.39 | 0.51 | **0.39** |
| DeepSeek | | 0.08 | 0.45 | 0.29 | 0.44 | 0.32 |
| DeepSeek+RAG | | **0.19** | 0.48 | 0.34 | 0.47 | **0.37** |

*4.3 Case-based Response Analysis*

This study empirically evaluates the domain knowledge enhancement effect of GraphRAG on general-purpose LLMs through comparative experiments. A comparison framework was established between DeepSeek models (baseline group) and DeepSeek+GraphRAG models (experimental group), utilizing a real-user query set of unstructured questions from "*Records of the Three Kingdoms*" which contain semantic ambiguities and colloquial expressions. The results demonstrate that GraphRAG integrates structured knowledge from knowledge graphs with the semantic understanding capabilities of large models, achieving multi-dimensional knowledge complementarity.

In the case study of "赤壁之战后，孙氏政权中哪些将领与周瑜存在师徒传承关系？" the baseline group identified only two core individuals. In contrast, the experimental group uncovered three additional indirectly related figures—Ling Tong (tactical succession), Zhu Ran (inheritance of defensive systems), and Lu Su (continuation of institutional frameworks)—and provided historical evidence and explanatory paths for these associations. This validates the technical advantages of the graph-enhanced framework in expanding knowledge coverage and constructing complex relational networks. Detailed findings are resented in Figure 10.

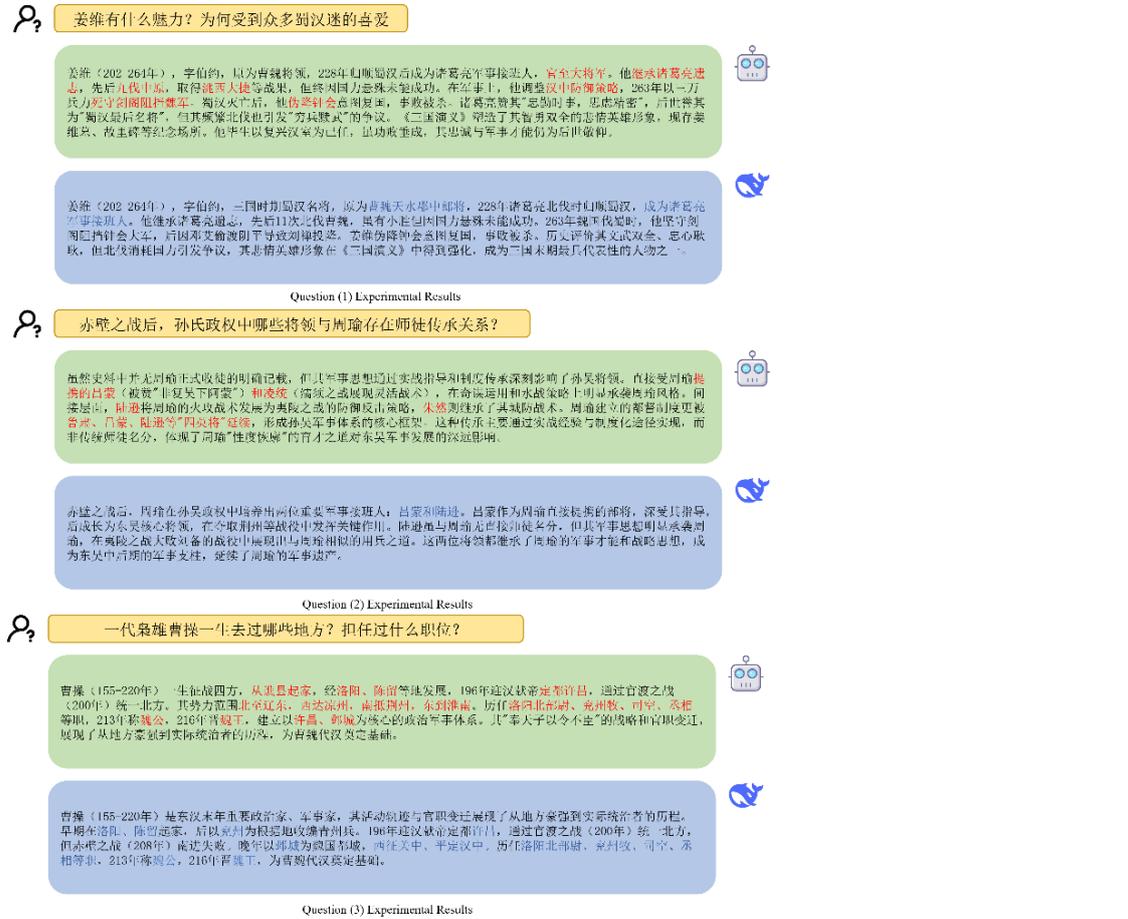

Figure 10. Comparative Case Study of Response Generation

## 5. Discussion and Conclusion

### 5.1 Discussion

In the process of constructing the knowledge graph, the Xunzi-Qwen1.5-14B model, enhanced with chain-of-thought (CoT) prompting, Simplified Chinese input and domain-specific fine-tuning, achieved optimal performance in relation extraction tasks (F1=0.68), significantly outperforming general-purpose models like DeepSeek (F1=0.23). This advantage can be attributed to two factors: first, CoT prompting effectively guides the model to capture deep semantic features of Classical Chinese by simulating human experts' step-by-step reasoning; second, the base model's continued pre-training and knowledge extraction fine-tuning on classical texts demonstrate the benefits of domain specialization. Additionally, process supervision techniques, through fine-grained control of reasoning paths via a scoring and reward model (as detailed in Table 7), enhance the precision and logical consistency of extraction results. This finding aligns with previous research on prompt engineering and supervision mechanisms (Liu et al. 2025), while our study uniquely combines CoT and process supervision to validate the feasibility of process supervision in low-resource classical text scenarios.

In the design of the GraphRAG system, the collaborative mechanism of graph-structured and semantic retrievals exhibits significant advantages. Experimental results show that integrating GraphRAG with the DeepSeek model improved the F1 score on the C-CLUE open-domain dataset by 11% (from 0.08 to 0.19), with a notable enhancement in the interpretability of generated results (as shown in Fig 10). This improvement arises from the knowledge graph's support for multi-hop reasoning and entity augmentation: the explicit associations between graph nodes enable the system to overcome traditional RAG's limitations in local semantic matching, uncovering indirect relational paths. Moreover, compared to the general-purpose GraphRAG framework proposed by Peng (Peng et al. 2024), our study introduces entity expansion retrieval optimization strategies tailored to the sparsity and entity ambiguity of historical texts, thereby enhancing the comprehensiveness and accuracy of query results.

*5.2 Conclusion*

This study proposes a GraphRAG construction method grounded in a historical-person knowledge graph, integrating the original "*The First Four Histories*" corpus, knowledge-graph techniques, and large language models to enhance general LLMs' extraction and retrieval performance on classical texts. Specifically, in knowledge-graph construction, experiments demonstrate that chain-of-thought prompting and Simplified Chinese input significantly boost model performance, yielding Xunzi-Qwen1.5-14B as the optimal model. Concurrently, a process-supervision–based scoring-and-reward model evaluates the extraction pipeline, achieving automatic relational-triple extraction from unstructured text. For graph-augmented retrieval generation, results show that the GraphRAG system—combining graph-structure retrieval with semantic-vector retrieval—outperforms standalone general LLMs, markedly improving historical knowledge extraction, multi-source data integration, and answer quality.

In summary, we conduct an in-depth analysis of the synergistic mechanisms between knowledge graphs and LLMs, focusing on two complementary paradigms: ①using knowledge graphs as external knowledge repositories to supply systematic semantic support for LLMs; and ②leveraging LLMs' intelligent processing capabilities to optimize the knowledge-graph construction workflow, substantially reducing manual annotation and extraction effort. This work extends the paradigm of Graph-based Retrieval-Augmented Generation in the domain of classical texts, delivering an efficient information-extraction framework for historical-person knowledge. By deeply integrating knowledge extraction, graph-structure retrieval, and generative question answering, the GraphRAG framework achieves dynamic, synergistic optimization beyond traditional isolated, staged pipelines.

This study has limitations. First, models such as Xunzi-Qwen1.5-14B require substantial computational resources, limiting their use in low-resource environments. Second, our corpus is confined to the Four Histories and may omit other classical texts; manual annotation and proofreading could introduce gaps, affecting the knowledge graph's comprehensiveness and generalizability. Future work will expand to additional historical sources, incorporate semi-automated annotation tools to improve dataset quality, and explore model-compression techniques—such as knowledge distillation—to lower the resource demands of domain-specific models.

**Funding**

This work was supported by the Jiangsu Provincial Social Science Foundation General Project of China [No.24TQB004].

## Data Availability

Not applicable.

**Table Legends**

Table 1 Sources of Historical Corpus
Table 2 Model Capabilities
Table 3 Key Hyperparameters for Model Fine-tuning
Table 4 Performance Evaluation of Prompt-Engineered
Table 5 Performance Evaluation of Input Models for Traditional and Simplified Chinese Characters
Table 6 Comparative Performance of General-Purpose and Domain-Specialized Models on Constrained Relation Extraction from Classical Texts
Table 7 Supervised Scoring Rules for Constrained-Domain Relation Extraction Across Reasoning Stages
Table 8 Exemplar Stepwise Scoring Process
Table 9 Performance Evaluation of Historical Figure Relationship GraphRAG System

**Figure Legends**

Figure 1 Research Framework
Figure 2 Modeling of Historical Character Relationships Knowledge
Figure 3 Example of Historical Corpus
Figure 4 Design of 5-Shot Prompt
Figure 5 Design of COT Prompt
Figure 6 Structure Illustration of the Instruction Dataset
Figure 7 Score Distribution Histogram for Knowledge Extraction Results
Figure 8 Exemplary Visualization of Historical Figure Knowledge Graph
Figure 9 System Workflow for Graph-based Retrieval-Augmented Generation with Historical Figure Context
Figure 10 Comparative Case Study of Response Generation